\documentclass[runningheads]{llncs}

\usepackage[english]{babel}

\usepackage[letterpaper,top=2cm,bottom=2cm,left=3cm,right=3cm,marginparwidth=1.75cm]{geometry}

\usepackage{multirow}
\usepackage{multicol}
\usepackage{tikz} 
\usepackage{sectsty}
\usepackage{amsmath}
\usepackage{graphicx}
\usepackage[colorlinks=true, allcolors=blue]{hyperref}
\usepackage{booktabs}
\usepackage{adjustbox}
\title{BLUEX: A benchmark based on Brazilian Leading Universities Entrance eXams}
\author{Thales Sales Almeida\inst{1,2} \and Thiago Laitz\inst{1,3} \and Giovana K. Bonás\inst{1} \and Rodrigo Nogueira\inst{1,2}}
\institute{State University of Campinas (UNICAMP) \and
Maritaca AI \and NeuralMind AI
}

\begin{document}
\maketitle


\begin{abstract}

One common trend in recent studies of language models (LMs) is the use of standardized tests for evaluation. However, despite being the fifth most spoken language worldwide, few such evaluations have been conducted in Portuguese. This is mainly due to the lack of high-quality datasets available to the community for carrying out evaluations in Portuguese. To address this gap, we introduce the Brazilian Leading Universities Entrance eXams (BLUEX), a dataset of entrance exams from the two leading universities in Brazil: UNICAMP and USP. The dataset includes annotated metadata for evaluating the performance of NLP models on a variety of subjects. Furthermore, BLUEX includes a collection of recently administered exams that are unlikely to be included in the training data of many popular LMs as of 2023. The dataset is also annotated to indicate the position of images in each question, providing a valuable resource for advancing the state-of-the-art in multimodal language understanding and reasoning. We describe the creation and characteristics of BLUEX and establish a benchmark through experiments with state-of-the-art LMs, demonstrating its potential for advancing the state-of-the-art in natural language understanding and reasoning in Portuguese. The data and relevant code can be found at \url{https://github.com/Portuguese-Benchmark-Datasets/BLUEX}
\end{abstract}

\section{Introduction}

Recent advances in Language Models (LMs) have generated significant interest due to their demonstrated capabilities on a wide range of language tasks, including text classification, language translation, and text generation~\cite{palm,chinchila}. LM performance has been particularly impressive on standardized tests, which present challenging questions requiring high levels of domain-specific knowledge and reasoning. For instance, recent benchmarks on GPT-4~\cite{openai2023gpt4} showed that it can achieve human-level performance on a variety of graduate-level benchmarks.

Despite the impressive performance of LMs on standardized tests, few evaluations have been performed in Portuguese~\cite{EnemPaper}, partially due to the lack of available datasets in the language. This lack of high-quality, standardized datasets presents a significant challenge for researchers interested in developing and evaluating LMs in Portuguese. To address this gap for Brazilian Portuguese, we introduce BLUEX, a dataset consisting of entrance exams for the two leading universities in Brazil. Our dataset offers a rich source of high-quality high school-level questions annotated with their respective subjects, as well as flags indicating the required capabilities necessary to respond accurately to the questions, such as knowledge of Brazilian culture and the application of mathematical reasoning. These annotations can be used to evaluate the performance of LMs on a variety of subjects and capabilities such as domain-specific knowledge and reasoning. Additionally, BLUEX includes a collection of recently administered entrance exams that are unlikely to be included in the training data of many currently popular LMs.

In anticipation of the emergence of multimodal models that combine text and image understanding, we have annotated BLUEX to indicate the position of images in each question. Additionally, we have included all necessary images with the dataset to facilitate research on multimodal language tasks. We believe that this resource will be essential in evaluating the performance of models that reason with both text and image inputs to solve complex problems.

In this paper, we describe the creation and characteristics of BLUEX and establish a benchmark through experiments with state-of-the-art LMs. Our findings suggest that BLUEX provides a valuable resource for benchmarking and advancing the state-of-the-art in natural language understanding and reasoning in Portuguese. This is particularly relevant since even the current state-of-the-art models, such as GPT-4, still have considerable room for improvement and do not achieve the highest cutoff grades for both universities.

\section{Related Work}

In the realm of Portuguese Natural Language Processing (NLP) datasets, there appears to be a limited availability.

For question-answering tasks, Faquad \cite{faquad} is available, which exhibits an extractive style akin to SQuAD~\cite{rajpurkar2016squad}. It features questions concerning Brazilian higher education institutions, with documents sourced from a federal university and supplemented by Wikipedia articles. Another option is the Multilingual Knowledge Questions and Answers (MKQA) dataset, which covers 26 languages~\cite{longpre2021mkqa}. This dataset was generated by selecting 10,000 queries from the Natural Questions dataset~\cite{kwiatkowski2019natural} and acquiring new passage-independent answers for each question. Subsequently, human translators translated the questions and answers into 25 non-English, typologically diverse languages, including Portuguese.

Regarding sentence entailment tasks, ASSIN 1 and 2~\cite{fonseca2016assin,real2020assin} are available. These datasets encompass Recognizing Textual Entailment (RTE), also referred to as Natural Language Inference (NLI), and Semantic Textual Similarity (STS) tasks. The former involves predicting if a given text (premise) implies another text (hypothesis), while the latter quantifies the semantic equivalence between two sentences.

The Portuguese Language Understanding Evaluation (PLUE) benchmark~\cite{Gomes2020} provides Portuguese translations of the GLUE~\cite{wang2018glue}, SNLI~\cite{bowman2015large}, and SciTAIL~\cite{khot2018scitail} datasets. These translations have been generated using automatic translation tools including Google Translate and OpusMT~\cite{TiedemannThottingal:EAMT2020}.

The Winograd Schema Challenge (WSC) dataset~\cite{kocijan2020review} contains pairs of sentences with minimal differences, featuring an ambiguous pronoun that is resolved divergently between the two sentences. Melo et al.~\cite{melo2019winograd} manually translated and adapted this dataset to Portuguese.

For sentiment analysis tasks, the TweetsentBr dataset \cite{tweetsentbr} consists of 15,000 tweets related to the TV show domain, collected between January and July 2017. The tweets were manually annotated by seven annotators into three classes: positive, neutral, and negative.

The Multilingual Amazon Slu resource package (SLURP) for Slot-filling, Intent classification, and Virtual assistant Evaluation (MASSIVE)~\cite{fitzgerald2022massive} is a 1M-example dataset containing realistic virtual utterances in 51 languages, including Portuguese. Professional translators translated the dataset from English, and it is annotated for slot (55 classes) and intent (60 classes) prediction tasks.

A dataset more closely related to BLUEX is the ENEM-challenge dataset \cite{enemChalengeIntro}, which includes the editions of the Brazilian national exam, Exame Nacional do Ensino Medio (ENEM), from 2009 to 2017. Additionally, Nunes et al.~\cite{EnemPaper} introduced a dataset containing the ENEM exam of 2022, the same paper evaluated the performance of LMs such as GPT-3.5-Turbo and GPT-4 on both the ENEM-challenge and the ENEM 2022 datasets.

\section{The BLUEX Dataset}

\subsection{Dataset Creation}

BLUEX is a dataset comprising more than 1,000 multiple choice questions from the entrance exams of the two leading universities in Brazil, Unicamp and USP, administered between 2018 and 2023. The dataset was created by automatically extracting each question text, alternatives, and related images using scripts, and subsequently each example was manually annotated to correct extraction errors and provide additional metadata such as image positioning.

\subsection{Annotated Question Metadata}

The annotated metadata is described below.
\begin{itemize}
    \item \textbf{Prior Knowledge (PRK)} - Indicates whether the question requires knowledge from outside of what has been provided in the question, such as familiarity with a particular author's work or a specific mathematical formula.
    \item \textbf{Text Understanding (TU)} - Indicates whether the question requires understanding of a particular text.
    \item \textbf{Image Understanding (IU)} - Indicates whether the question requires understanding of an image. It should be noted that not all questions with images require their understanding to answer the question.
    \item \textbf{Mathematical Reasoning (MR)} - Indicates whether the question requires mathematical reasoning, such as the ability to perform calculations and symbolic manipulations.
    \item \textbf{Multilingual (ML)} - Indicates whether the question requires knowledge of two or more languages, such as questions designed to test English skills of Portuguese speakers.
    \item \textbf{Brazilian Knowledge (BK)} - Indicates whether the question involves knowledge specific to Brazil, such as Brazilian history, literature, geography, or culture.
    \item \textbf{Subjects} - A list of subjects related to the question, such as geography, physics, etc.
    \item \textbf{Related Images} - A list of all the related images for the question.
    \item \textbf{Alternative Type} - Indicates whether the answer choices are presented as text or as images. This is important because some questions may use images as answer choices, which requires different processing techniques than questions with only textual answers.
\end{itemize}

By providing such annotations along with the questions we aim to facilitate research into language understanding and reasoning in Portuguese for both pure language models and multimodal models. We believe that BLUEX will be a valuable resource for researchers to evaluate and improve the performance of future language models in the context of Portuguese-language standardized tests.

\subsection{Image Positioning}

\begin{figure}[!htb]
    \centering
    \includegraphics[width=0.8\columnwidth]{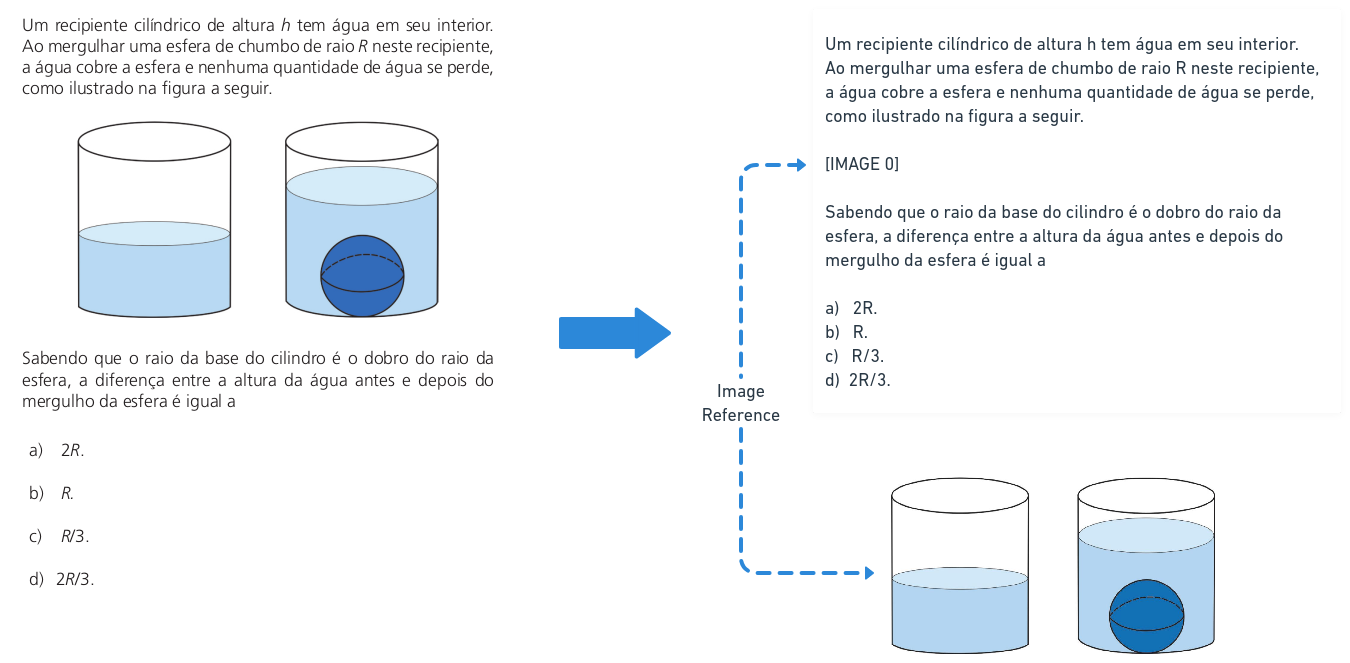}
    \caption{Example of image annotation in BLUEX.}
    \label{fig:image_anot_example}
\end{figure}

Many of the questions in the exams require a contextual or informational understanding of images. Despite active research in the field of multimodal models, models that can adeptly process both text and image data and yield satisfactory results remain scarce in the public domain. We believe that BLUEX can serve as an essential evaluation tool for such models. 
Anticipating the use of models that will process images and text in an interleaved manner, we also provide precise information regarding the placement of images within the question, as illustrated in Figure \ref{fig:image_anot_example}.

\subsection{Dataset Distribution}

The BLUEX dataset covers a wide range of high school subjects, including Mathematics, Physics, Chemistry, Biology, History, Geography, English, Philosophy and Portuguese, as well as multidisciplinary questions that involve two or more subjects. The distribution of questions is shown in Table \ref{tab:per_subjetc}, where we also provide the distribution for the subset of questions without images, which accounts for approximately 58\% of the total dataset.

Furthermore, Table \ref{tab:per_category} shows the distribution of the dataset across annotated categories, as explained in Section 3.2. We observe that the majority of questions require specific knowledge and the ability to comprehend text, two expected capabilities in students taking these exams. Note that any given question can be part of multiple categories.

\begin{table}[]
\centering
\begin{adjustbox}{width=1\textwidth}
\begin{tabular}{@{}cccccccccccc@{}}
\toprule
                 & biology & chemistry & english & geography & history & mathematics & philosophy & physics & portuguese & multidisciplinary & Total \\ \midrule
\textbf{UNICAMP}          & 60      & 45        & 57      & 51        & 60      & 89          & 1          & 61      & 86         & 46                & 556   \\
\textbf{USP}              & 50      & 63        & 41      & 55        & 63      & 69          & 4          & 63      & 88         & 43                & 539   \\
\textbf{BLUEX}            & 110     & 108       & 98      & 106       & 123     & 158         & 5          & 124     & 174        & 89                & 1095  \\ \midrule
\multicolumn{12}{c}{\textbf{No images}}                                                                                                              \\ \midrule
\textbf{UNICAMP} & 35      & 15        & 20      & 22        & 49      & 64          & 1          & 36      & 65         & 31                & 338   \\
\textbf{USP}     & 23      & 15        & 25      & 16        & 52      & 36          & 4          & 26      & 80         & 23                & 300   \\
\textbf{BLUEX}   & 58      & 30        & 45      & 38        & 101     & 100         & 5          & 62      & 145        & 54                & 638   \\ \bottomrule
\end{tabular}
\end{adjustbox}
\caption{Distribution over subjects.}
\label{tab:per_subjetc}
\end{table}

\begin{table}[]
\centering
\begin{adjustbox}{width=0.4\textwidth}

\begin{tabular}{@{}lcccccc@{}}
\toprule

                              & DS  & TU  & IU  & MR  & ML  & BK  \\ \midrule
\textbf{UNICAMP}                        & 431 & 440 & 160 & 209 & 60  & 69  \\
\textbf{USP}                           & 446 & 442 & 203 & 174 & 43  & 63  \\
\textbf{BLUEX}          & 877 & 882 & 363 & 383 & 103 & 132 \\ \midrule
\multicolumn{7}{c}{No Images}                                     \\ \midrule
\textbf{UNICAMP}              & 273 & 282 & 0   & 118 & 23  & 46  \\
\textbf{USP}                  & 237 & 269 & 0   & 70  & 25  & 43  \\
\textbf{BLUEX} & 510 & 551 & 0   & 188 & 48  & 89  \\ \bottomrule
\end{tabular}
\end{adjustbox}
\caption{Distribution over categories.}
\label{tab:per_category}
\end{table}


\section{Results}

\begin{table}[]
\centering
\begin{adjustbox}{width=0.7\textwidth}
\begin{tabular}{l|ccc|cc}
\toprule
Model                                                & BLUEX                        & UNICAMP                         & USP                             & MR                              & BK                              \\ \midrule
Highest Cutoff Score                                 & 0.863                           & 0.855                           & 0.872                           &   -                              &               -                  \\
Average Human Score                                  & 0.521                           & 0.530                           & 0.511                           &    -                            &    -                           \\
Random                                               & 0.220                           & 0.250        & 0.200       & 0.223                           & 0.228                           \\ \midrule
GPT-4 \cite{openai2023gpt4}                                                & \textbf{0.748} & \textbf{0.749} & \textbf{0.747} & \textbf{0.447} & \textbf{0.854} \\
Sabiá 65B \cite{pires2023sabia}     & 0.632                           & 0.615                           & 0.650                           & 0.239                           & 0.775                           \\
GPT-3.5-Turbo                                              & 0.582                           & 0.580                           & 0.583                           & 0.277                           & 0.764                           \\
LLaMA 65B \cite{touvron2023LLaMA}   & 0.542                           & 0.530                           & 0.557                           & 0.271                           & 0.652                           \\
OPT 66B \cite{zhang2022opt}         & 0.223                           & 0.246                           & 0.197                           & 0.186                           & 0.258                           \\ \midrule
Sabiá 7B \cite{pires2023sabia}      & \textbf{0.466} & \textbf{0.494} & \textbf{0.433} & 0.25                            & \textbf{0.551} \\
Alpaca 7B \cite{alpaca}                                           & 0.284                           & 0.308                           & 0.257                           & \textbf{0.261} & 0.258                           \\
BloomZ 7B \cite{bloomz}             & 0.284                           & 0.275                           & 0.293                           & 0.17                            & 0.326                           \\
LLaMA 7B \cite{touvron2023LLaMA}    & 0.255                           & 0.275                           & 0.233                           & 0.255                           & 0.247                           \\
Bertin 6B \cite{delarosa2022bertin} & 0.241                           & 0.293                           & 0.183                           & \textbf{0.261} & 0.315                           \\
Bloom 7B \cite{workshop2023bloom}   & 0.238                           & 0.302                           & 0.167                           & 0.255                           & 0.281                           \\
XGLM 7.5B \cite{xglm}                 & 0.205                           & 0.219                           & 0.19                            & 0.213                           & 0.202                           \\
OPT 6.7B \cite{zhang2022opt}          & 0.205                           & 0.240                           & 0.167                           & 0.207                           & 0.281                           \\
GPT-J 6B \cite{gpt-j}                                                & 0.197                           & 0.222                           & 0.17                            & 0.186                           & 0.236                           \\ \bottomrule
\end{tabular}
\end{adjustbox}

\caption{Accuracy in the BLUEX dataset.}
\label{tab:main_results}
\end{table}

To enable future comparisons, we evaluated our dataset using several language models, ranging from 6B to 66B parameters, including OpenAI's GPT-4 and GPT-3.5-Turbo models. Our experiments were conducted using large language models with no specific training for this task. Each model was provided with one example in the input and then asked to answer a question from the test set. The example was randomly selected from an exam of the same university as the current question, but from a different year. For example, if the current question is from UNICAMP 2019, the example provided in the prompt would be a question from a UNICAMP exam, but not from 2019. We excluded all questions containing images from our experiments since the language models we used can only process text. This resulted in a total of 638 questions being used, which corresponds to approximately 60\% of the dataset

Table~\ref{tab:main_results} summarizes our experimental findings, including the mean score achieved by exam-taking students, as well as the mean cutoff score of the most competitive major, which is medicine in both universities.\footnote{The average and cutoff scores are reported by the entities responsible for administering the exams. The results presented in Table~\ref{tab:main_results} are the average of all the exams contained in the BLUEX dataset.} The BLUEX column shows the accuracy of the whole subset used in the evaluation, while the UNICAMP and USP columns account for only the questions from the respective universities. The MR and BK columns account only for questions that include those categories.

Among the language models tested in the 7B-parameter range, Sabiá~\cite{pires2023sabia}, a model further pre-trained in Portuguese, consistently outperformed all other models, coming close to matching the average human score. Among the open-source models in the 60B-parameter range, LLaMA 65B~\cite{touvron2023LLaMA} significantly outperformed OPT 66B~\cite{zhang2022opt} and achieved similar performance to GPT-3.5-Turbo. Sabiá 65B achieved better performance than GPT-3.5-Turbo but still lagged behind GPT-4 by ten points. GPT-4 was by far the best model in our evaluations but did not achieve an average score high enough to pass in medicine, the most competitive major. It is worth noting that the average and cutoff scores provided in Table~\ref{tab:main_results} are computed taking into account the whole exam, including questions with images, while the scores obtained by the language models utilize only the subset of questions with no images.

We also conducted a more detailed analysis of the models' performance by examining their ability to handle specific question types. Table~\ref{tab:main_results} presents the findings for questions that required Mathematical Reasoning (MR) and Brazilian Knowledge (BK). We observe that, with the exception of GPT-4, all models struggled to perform significantly better than random chance in questions that required Mathematical Reasoning. Even GPT-4 only achieved an accuracy of 44\% in MR questions. On the other hand, when considering questions that require brazilian knowledge, Sabiá greatly outperformed all the other models in the 7B-parameter range, indicating that the extra pretraining in Portuguese provided the model with additional regional knowledge. In the 60B-parameter range, Sabiá also showed improvement over LLaMA, increasing the accuracy in these questions by 10 points and slightly outperforming GPT-3.5-Turbo. Nevertheless, it could not match the remarkable performance of GPT-4.

\begin{figure}[!htb]
    \centering
    \includegraphics[width=0.8\columnwidth]{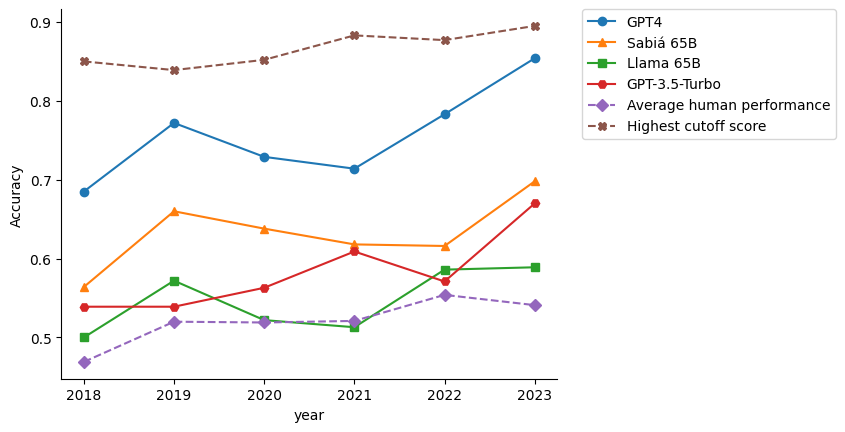}
    \caption{Accuracy of the best models over the years of the exams.}
    \label{fig:performance_over_years}
\end{figure}

Moreover, Figure \ref{fig:performance_over_years} displays the performance of the top four models on the exams conducted each year. It can be observed that the models have a small variance between the years, which is expected as the difficulty of each exam and the number of questions in the subset vary across years. A surprising result, however, is the increased performance that all models seem to exhibit in 2023. The average and highest cutoff scores also increased slightly over the years, indicating that the exams became slightly easier in recent years. Since the 2023 exams were very recently administered, it is unlikely that they are part of any of the studied models' training data. Therefore, since the models' performance in the most recent years is comparable to that in older exams, it is reasonable to assume that the models are not merely memorizing the answers for the questions in the dataset.

\section{Conclusion}

This work introduced BLUEX, a new dataset that consists of 13 college entrance exams applied between 2018 and 2023 from two of the leading Brazilian universities, UNICAMP and USP. Each question of these exams was extensively annotated to help measure different abilities across multiple subjects in Portuguese. Beyond that, by providing images and their corresponding positions within the text, BLUEX is one of the few Portuguese datasets that are ready to evaluate multimodal models. We provide results from multiple LMs as baselines and reference scores based on students performance to facilitate future comparisons. We believe that BLUEX will be a important benchmark in the evaluation of the Portuguese capabilities of future models.

\section{Future Work}

The models used in this study employed a single in-context example. However, there's room for further investigation, such as determining whether increasing the number of few-shot examples could boost the performance of each model, as well as assessing their zero-shot performance. Furthermore, Nunes et al.~\cite{EnemPaper} showed that GPT-4's performance on ENEM questions was significantly boosted when chain-of-thought prompts~\cite{wei2022chain} were used. Adopting a similar approach here could potentially lead to performance improvement.

Finally, regarding multimodal models, their performance can be assessed utilizing the BLUEX dataset. This provides an opportunity for researchers to investigate the models' capabilities in integrating visual and textual information to address high school level questions.

\bibliographystyle{splncs04}
\bibliography{sample}

\section{Appendix}

\subsection{Prompt for evaluation}

\begin{figure}[!htb]
\sffamily
\tiny
\centering
\begin{tikzpicture} 
\node (table) [inner sep=2pt] { 
\begin{tabular}{p{0.001\columnwidth}p{0.06\columnwidth}p{0.85\columnwidth}}

\multirow{24}{*}{\rotatebox{90}{}}
& \multicolumn{2}{l}{{\color[HTML]{343434} \textbf{Select the correct alternative}}}                      \\ \\ \\
& \multicolumn{2}{l}{{\color[HTML]{343434} \textit{... Few-shot examples go here, separated with \#\#\#}}} \\ \\
& \multicolumn{2}{l}{{\color[HTML]{343434} \textbf{Nth example:}}}                      \\
& {\color[HTML]{343434} \textbf{Question:}} & {\color[HTML]{343434} Times change, desires change, Beings change, trust changes: The whole world is composed of change, Always taking on new qualities. Continually we see novelties, Different in everything from hope: From evil, only the sorrows remain in memory, And from good (if any existed), the longing. Time covers the ground with a green cloak, Which was once covered in cold snow, And in me, it turns the sweet song into tears. And besides this daily change, Another change causes even more astonishment, That no longer changes as it used to. (Luís Vaz de Camões). (Luís de Camões, 20 sonnets. Campinas: Unicamp Publisher, p.91.) Indicate the statement that applies to the sonnet written by Camões.} \\
& \multirow[t]{4}{*}{{\color[HTML]{343434} \textbf{Alternatives:}}} & {\color[HTML]{343434} A. The poem takes up the Renaissance theme of the change of things, which the poet feels as a reason for hope and faith in life.} \\ 
&                                                                   & {\color[HTML]{343434} B. The idea of transformation refers to worldly things, but it does not affect the poet's state of mind due to his love belief.} \\ 
&                                                                   & {\color[HTML]{343434} C. Everything is always renewed, unlike the poet's hopes, which harbor his sorrows and longings.} \\ 
&                                                                   & {\color[HTML]{343434} D. Not only does the poet's state of mind change, but also his experience of change itself.} \\ 
& {\color[HTML]{343434} \textbf{Answer:}}   & {\color[HTML]{FF0000} \textit{D.}}        

\end{tabular}
}; 
\draw [rounded corners=.5em] (table.north west) rectangle (table.south east); 
\end{tikzpicture} 
\caption{Example of prompt used in the experiments, the question was translated into English for the convenience of readers. The text in red is the expected output.}
\label{fig:example_prompts}
\end{figure}
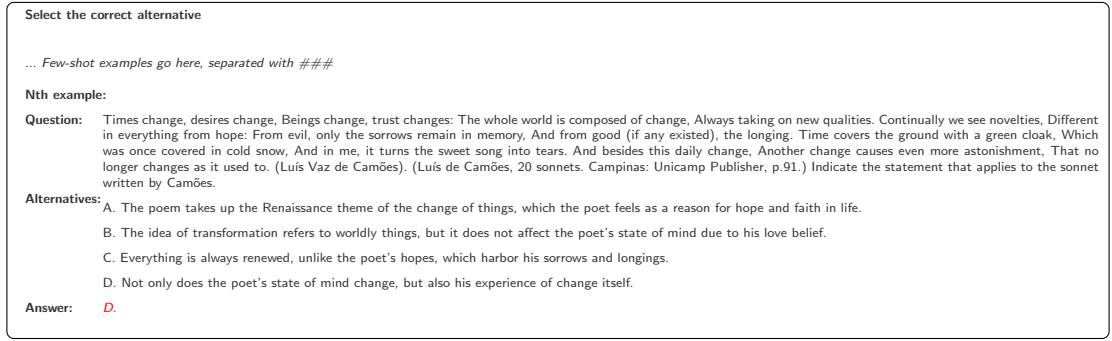

The prompt used for all the experiments in this paper is shown in the Figure \ref{fig:example_prompts}.

\subsection{Benchmark per Subject}

Table \ref{tab:full_results_experiment_per_subject} provides a detailed report of each model achieved accuracy by subject. Questions that were associated with more than one subject contributed to the accuracy of both scores. For example, a question related to mathematics and English will be taken into account when calculating the accuracy of both mathematics and English subjects.

\begin{table}[]
\centering
\begin{adjustbox}{width=1\textwidth}
\begin{tabular}{@{}l|ccccccccc@{}}
\toprule
Model     & Biology & Chemistry & English & Geography & History & Mathematics & Philosophy & Physics & Portuguese \\ \midrule
GPT-4      & 0.871   & 0.675     & 0.918   & 0.935     & 0.930    & 0.389       & 1.000        & 0.557   & 0.805      \\
Sabiá 65B & 0.771   & 0.350      & 0.837   & 0.774     & 0.883   & 0.278       & 1.000        & 0.257   & 0.755      \\
GPT-3.5-Turbo   & 0.700     & 0.350      & 0.714   & 0.806     & 0.805   & 0.259       & 0.714      & 0.329   & 0.629      \\
LLaMA 65B & 0.657   & 0.350      & 0.816   & 0.677     & 0.719   & 0.306       & 0.429      & 0.286   & 0.572      \\
OPT 66B   & 0.229   & 0.275     & 0.286   & 0.161     & 0.273   & 0.176       & 0.286      & 0.200     & 0.189      \\ \midrule
Sabiá 7B  & 0.514   & 0.350      & 0.592   & 0.565     & 0.672   & 0.241       & 0.571      & 0.271   & 0.509      \\
Alpaca 7B & 0.286   & 0.225     & 0.347   & 0.306     & 0.320    & 0.269       & 0.143      & 0.229   & 0.264      \\
BloomZ 7B & 0.243   & 0.075     & 0.551   & 0.371     & 0.336   & 0.185       & 0.143      & 0.171   & 0.308      \\
LLaMA 7B  & 0.229   & 0.325     & 0.286   & 0.210      & 0.266   & 0.231       & 0.000        & 0.314   & 0.245      \\
Bertin 6B    & 0.186   & 0.225     & 0.347   & 0.226     & 0.234   & 0.259       & 0.286      & 0.243   & 0.245      \\
Bloom 7B  & 0.243   & 0.225     & 0.327   & 0.210      & 0.219   & 0.259       & 0.143      & 0.214   & 0.239      \\
XGLM 7.5B   & 0.143   & 0.300       & 0.245   & 0.161     & 0.164   & 0.204       & 0.000        & 0.171   & 0.264      \\
OPT 6.7B    & 0.186   & 0.250      & 0.143   & 0.145     & 0.234   & 0.185       & 0.000        & 0.257   & 0.214      \\
GPTJ 6B     & 0.214   & 0.200       & 0.204   & 0.113     & 0.227   & 0.194       & 0.000        & 0.200     & 0.195      \\ \bottomrule
\end{tabular}
\end{adjustbox}
\caption{Results for each model by subject in BLUEX.}
\label{tab:full_results_experiment_per_subject}
\end{table}
\end{document}